Title:

# Feature Aggregation for Efficient Continual Learning of Complex Facial Expressions

Authors:

Thibault Geoffroy, thibault.geoffroy@esiea.fr [1]

Myriam Maumy, myriam.maumy@ehesp.fr [2]

Lionel Prevost, lionel.prevost@esiea.fr [1, 3]

Affiliations:

[1]: Learning Data Robotics (LDR) ESIEA Lab, ESIEA Paris, 75005 Paris

[2] : Laboratoire Arènes, UMR CNRS 6051, équipe RSMS EHESP, 2-10, rue D’Oradour-sur-Glane • 75015 PARIS

[3] : CRI, Univ. Paris 1 Panthéon-Sorbonne, 90 rue de Tolbiac, 75013 Paris

Abstract:

As artificial intelligence (AI) systems become increasingly embedded in our daily life, the ability to recognize and adapt to human emotions is essential for effective human–computer interaction. Facial expression recognition (FER) provides a primary channel for inferring affective states, but the dynamic and culturally nuanced nature of emotions requires models that can learn continuously without forgetting prior knowledge. In this work, we propose a hybrid framework for FER in a continual learning setting that mitigates catastrophic forgetting. Our approach integrates two complementary modalities: deep convolutional features and facial Action Units (AUs) derived from the Facial Action Coding System (FACS). The combined representation is modelled through Bayesian Gaussian Mixture Models (BGMMs), which provide a lightweight, probabilistic solution that avoids retraining while offering strong discriminative power. Using the Compound Facial Expression of Emotion (CFEE) dataset, we show that our model can first learn basic expressions and then progressively recognize compound expressions. Experiments demonstrate improved accuracy, stronger knowledge retention, and reduced forgetting. This framework contributes to the development of emotionally intelligent AI systems with applications in education, healthcare, and adaptive user interfaces.

# 1 Introduction

Advances in Artificial Intelligence (AI) have made intelligent systems an integral part of our daily lives. With increasing integration into our technological environment, the ability of AI to assess and distinguish between different emotional states has become essential for improving human–computer interaction (Cowie et al., 2001; Picard, 2000). Facial Expression Recognition (FER) provides a channel to infer, analyze, and study how individuals express their emotional states through facial expressions.

The most common categorization of emotions is Ekman's basic emotions (Ekman and Friesen, 1971), which consist of six basic categories: anger, fear, disgust, happiness, sadness, and surprise (see Fig. 1), along with the neutral state representing the absence of emotion. This representation has become a baseline for emotion analysis. Its simplicity and universality have enabled the creation of numerous datasets in the FER context, and most FER algorithms rely on Ekman's annotation as a basis for classification. While this categorization is useful, in real-life contexts it can be limiting due to the complexity and subtlety of emotions. For this reason, other annotation systems were developed to analyze emotions in specific contexts such as learning. For example, Arthur Graesser and Sidney D'Mello (Graesser and D'Mello, 2012) uses a set of annotations for emotions occurring during the learning of difficult material. Such context-specific annotations can be very useful to study emotional states in activities like learning, job interviews, or clinical settings. However, creating datasets for such subtle expressions is both complex and time-consuming, which also complicates the development of FER systems for their classification. An alternative approach is to study compound expressions (Fig. 2) which represent facial expression formed by the combination of two or more basic emotion. These expressions are not represented by a simple combined activations of the facial muscles from the separate expressions but are considered distinct and recognizable affective state which allow us to move beyond the basic expressions and study more nuanced and complex expressions which are more representative of real-life interactions.

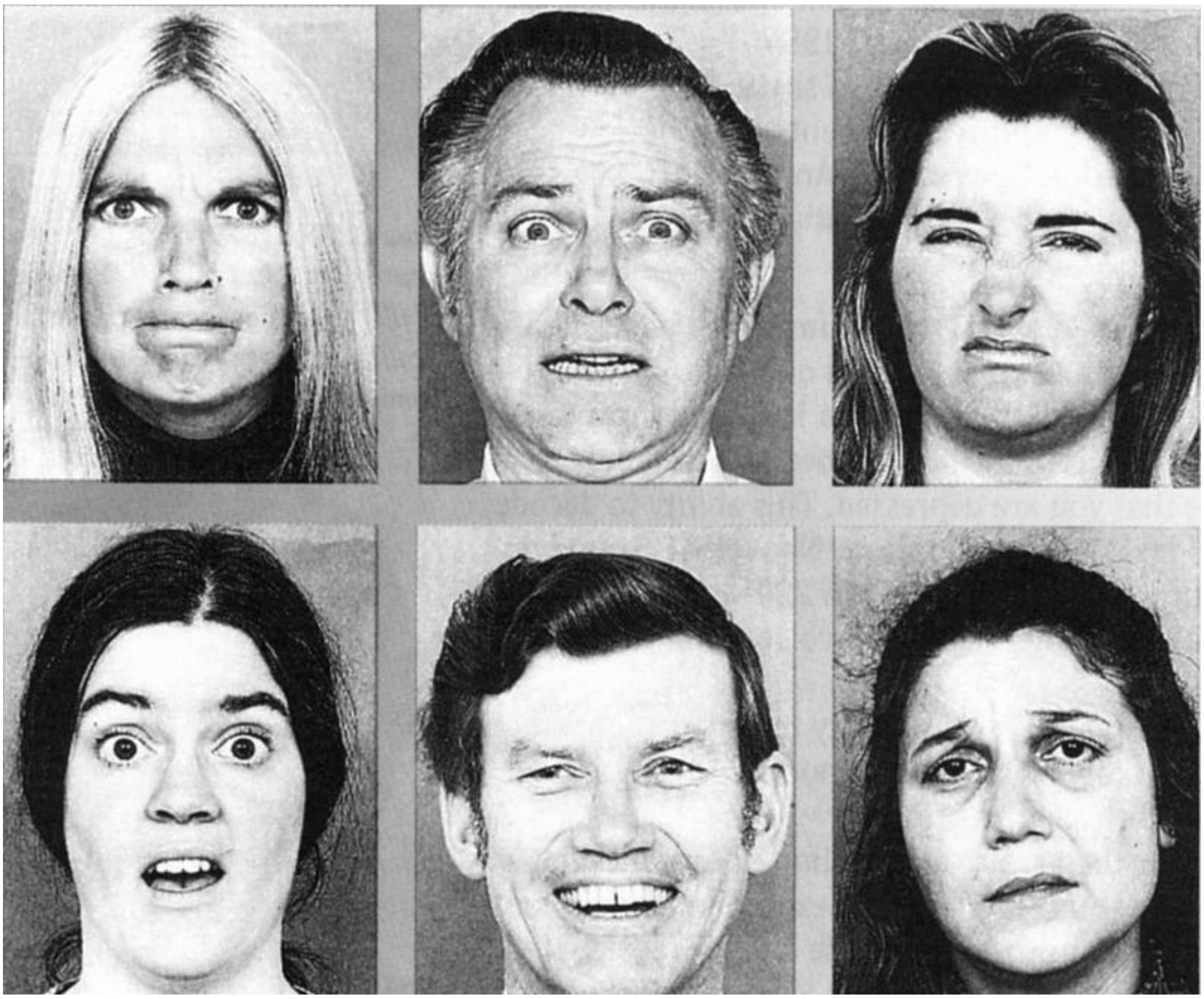

*Fig. 1. Ekman's 6 emotional expressions (Geslin, 2013).*

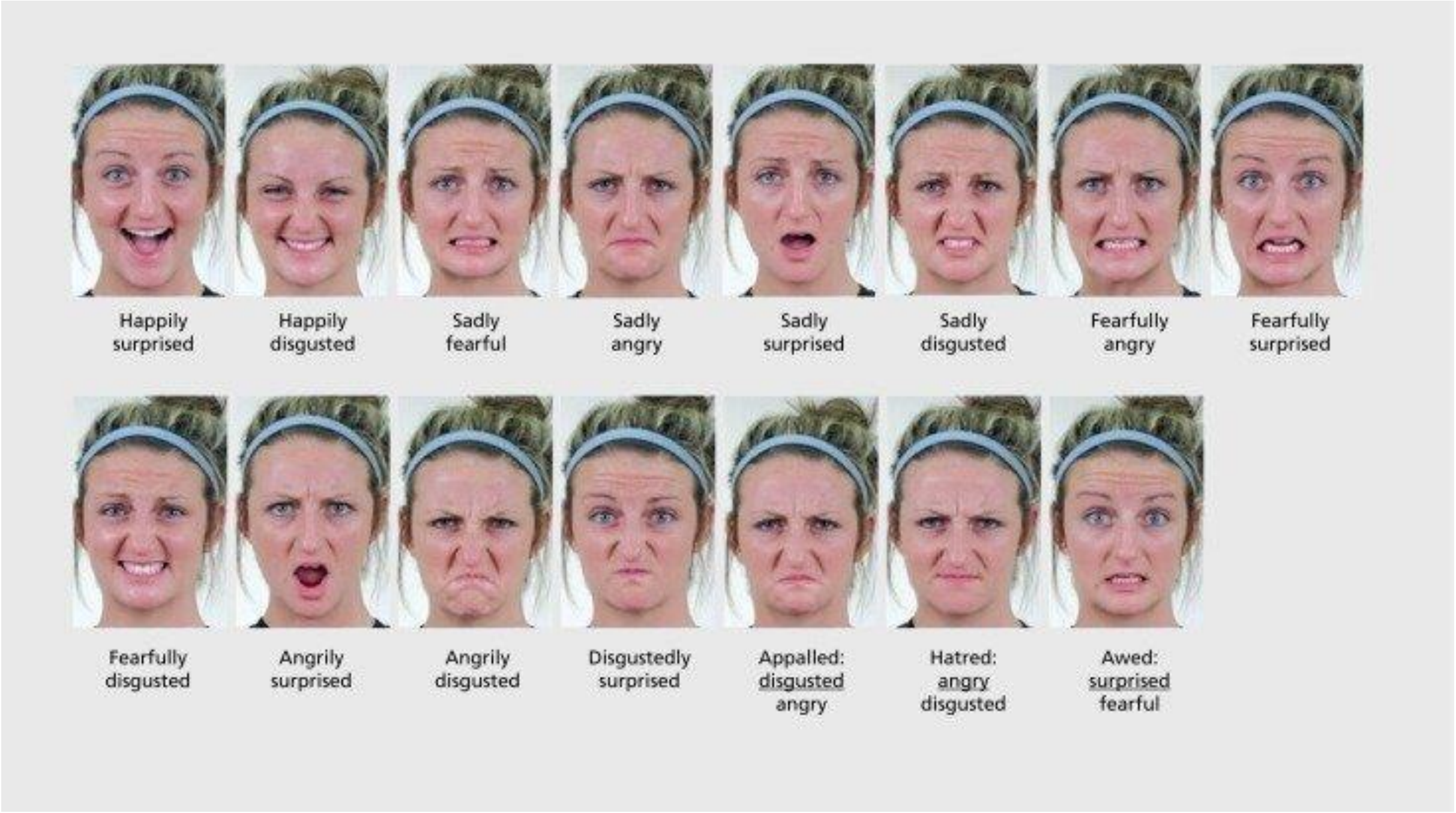

*Fig. 2 Compound Facial Expressions (Du et al., 2014).*

In addition to end-to-end systems such as Convolutional Neural Networks (CNNs), which learn features jointly with expression classification, other methods can be used to extract features for FER. Facial Action Units (AUs), introduced by Ekman and Friesen (Ekman and Friesen, 1978) , categorize facial muscle activations, and provide a structured and objective representation of facial expressions. Integrating multiple feature types for classification has already proven effective (Castellano et al., 2008), particularly in scenarios with limited training data, where continual learning is especially challenging. Although CNN-extracted features are powerful for the tasks they are trained on, their generalization is reduced on unseen tasks, and continual training of CNNs can increase catastrophic forgetting(Kirkpatrick et al., 2017).

Continual learning is a machine learning paradigm that enables models to learn continuously during deployment. In this setting, data become available sequentially, either one instance at a time or in batches, and the model is trained with little or no access to previous data. Without appropriate strategies, this leads to significant knowledge loss on previously learned tasks, a phenomenon known as catastrophic forgetting. This issue is one of the main challenges in continual learning and has been extensively studied (French, 1999; Kirkpatrick et al., 2017; Robins, 1995). The objective of continual learning is to acquire new information while retaining old knowledge. Some methods address this by storing parts of the dataset, but this increases memory requirements and raises privacy concerns. For this reason, example-free methods are increasingly studied (Chaudhry et al., 2019; Lopez-Paz and Ranzato, 2017).

Many applications of continual learning involve learning new classes of objects. Datasets such as CIFAR-100 (Krizhevsky and Hinton, 2009) and ImageNet (Deng et al., 2009) are benchmarks for such tasks, where the model must recognize very different objects, ranging from apples to cars or even porcupines. These diverse recognition tasks impose high complexity, as the model must make accurate predictions without suffering catastrophic forgetting. In the FER context, however, continual learning tasks are not completely disjoint: all data consist of faces, and the goal is to detect and classify new emotional expressions. For example, the Compound Facial Expressions of Emotion (CFEE) dataset (Du et al., 2014) contains both Ekman’s basic emotions and compound expressions. These compound expressions are distinct and culturally consistent facial patterns formed by the combination of two or more basic emotions, such as “happily surprised” or “sadly fearful.”

Several approaches have been proposed for such tasks, including regularization, and retraining of the model, but they perform poorly in this context. Moreover, since compound expressions are less frequent, the available labeled data are much more limited than for basic expressions, making neural network retraining difficult.

Since Action Units alone are not sufficient to classify compound expressions, and continuously retraining CNN feature extractors leads to catastrophic forgetting, our approach combines low-level features from a pre-trained CNN (Howard et al., 2017) with high-level features derived from Action Units. This hybrid representation enhances accuracy in compound expression recognition.

For the inference itself, most state-of-the-art FER approaches rely on deep neural networks, but these models are not always appropriate for continual learning. Probabilistic models offer several advantages: their ability to model class-conditional distributions allows class-by-class analysis and reduces vulnerability to catastrophic forgetting. Gaussian Mixture Models (GMMs) have shown strong performance in modeling data distributions, as FER classes often form clusters in the feature space. However, classical GMMs require fixing the number of mixture components in advance, which is a costly hyperparameter to optimize. To address this limitation, we employ Bayesian Gaussian Mixture Models (BGMMs), which automatically infer the optimal number of mixture components. This enables the model to adapt its complexity to the data and represent class distributions more robustly.

Our approach combines multimodal feature vectors with Bayesian Gaussian Mixture Models to achieve strong performance in facial expression recognition within a continual learning framework.

The two main contributions of this work are:

- Demonstrating the utility of multimodal feature fusion for accurately classifying both basic and compound facial expressions.
- Introducing an efficient probabilistic model that enables continual learning with minimal training overhead while effectively mitigating catastrophic forgetting compared to CNN-based approaches.

## 1.1 Related works

**Facial Expression Recognition**

Facial Expression Recognition (FER) is a pivotal domain in affective computing, aimed at automatically identifying human emotional expressions through facial cues. Traditional FER systems have historically relied on geometric or appearance-based features such as Active Shape Models (ASM) (Cootes et al., 1995), Local Binary Patterns (LBP) (Zhang et al., 1998), or Gabor filters (Abhishree et al., 2015). Though robust, these methods struggle with generalization when applied to unconstrained data (Kopalidis et al., 2024).

With the rise of deep learning, Convolutional Neural Networks (CNNs) (Alzubaidi et al., 2021) and their extensions such as Long Short-Term Memory (LSTM) models (Hochreiter and Schmidhuber, 1997) have significantly boosted FER performance, especially on datasets such as CK+ (Lucey et al., 2010), FER2013 (Goodfellow et al., 2013), or RAF-DB (Li et al., 2017). These models are often fine-tuned from pretrained image classification architectures (e.g., VGG, ResNet, EfficientNet) and consistently outperform classical methods by learning discriminative features in an end-to-end fashion (Kopalidis et al., 2024).

However, these deep learning approaches are typically trained in classical setting where the whole set of training data arrives all at once and are prone to catastrophic forgetting when required to adapt to new distribution of data or new classes. This makes them not suitable for continual learning.

**Continual Learning**

Continual learning has emerged as a necessary framework for machine learning models that must adapt over time to new data distributions and classes without forgetting previously acquired knowledge. The field categorizes approaches into regularization-based methods (EWC (Kirkpatrick et al., 2017)), replay-based strategies (FearNet (Kemker and Kanan, 2018) ), architecture-based expansions (ZOO (Ramesh and Chaudhari, 2022)) representation-based (DualNet (Pham et al., 2021) ) or optimization-based methods (GEM (Lopez-Paz and Ranzato, 2017)).

Unlike generic object classification benchmarks such as CIFAR-100 or ImageNet (Deng et al., 2009) FER tasks involve overlapping domains (all tasks concern facial data) but shifting label semantics (e.g., from basic to compound expressions). This makes continual learning in FER a more structured and domain-specific challenge. The Compound Facial Expression of Emotion (CFEE) dataset which includes both basic and compound expressions, is particularly well suited for investigating these dynamics thanks to its natural separation between simple and more complex emotions.

Yet, most continual learning methods are evaluated on object recognition datasets with clearly distinct tasks and classes. In contrast, FER involves different types of expressions derived from the same facial images, meaning that all tasks share a common basis of recognizing facial features. This strong similarity between tasks calls for approaches that can better exploit the inherent structure of FER data while remaining computationally efficient.

**Generative and Probabilistic Models**

Generative models such as GANs offer a solution to catastrophic forgetting by approximating past data distributions, enabling the virtual replay of old task data. However, these methods can be resource-intensive and remain prone to forgetting when continuously trained (Wang et al., 2024). An alternative lies in probabilistic models, which explicitly model data distributions. By modeling each class distribution separately, they mitigate catastrophic forgetting by preventing interference between new and old classes.

Among these, Gaussian Mixture Models (GMMs) (Dempster et al., 1977) and Bayesian Gaussian Mixture Models (BGMMs) (Corduneanu and Bishop, 2001) have demonstrated strong discriminative capabilities in FER (Tariq et al., 2013). In FER, where class distributions are naturally clustered yet sometimes highly overlapping (particularly for compound emotions), GMM-based approaches provide a lightweight and interpretable solution. Our work integrates BGMMs trained conditionally on each class within a multimodal representation space (combining CNN-extracted features and Action Units), enabling continual acquisition of knowledge on increasingly complex expressions while mitigating catastrophic forgetting.

Prior FER studies applying GMM or BGMM have mostly done so in classical settings and typically rely on a single feature modality, limiting their capabilities for complex tasks like recognizing compound expressions. Our work addresses these limitations by using BGMMs with several feature vectors extracted from the images combining CNN-extracted features and Action Units. This approach uses both low-level and high-level information, enabling more resilient continual learning and reducing catastrophic forgetting through the continual learning process.

# 2 Main Content

## 2.1 Method

### 2.1.1 Intuition

The main idea behind our model is the following: since we are working on a set of related tasks centered on facial image analysis, we chose to specialize the model within this specific domain rather than extend it to unrelated tasks. In this context, combining different types of features, with high-level features such as Action Units and low-level features extracted by a CNN-based encoder, enables the model to classify not only basic expressions but also more subtle ones, including compound expressions.

### 2.1.2 Architecture

Our model consists of three parts: an ensemble of feature extractors, a feature aggregator, and a probabilistic classifier. The feature extractors can be of different types (CNN-based features, Action Units, etc.) and are designed to compute complementary information, so that one can help compensate for the other's errors. In our implementation, we employ two feature extractors. The first is a CNN trained exclusively on basic emotions, from which we use the features extracted at the penultimate layer. The second is an Action Units extractor (Baltrusaitis et al., 2018). These two sources of information, one providing a low-level feature vector trained on the dataset and the other a high-level feature vector, together form a strong representation of the subject's face and emotional expression.

The second component of the model is a feature aggregator that combines the outputs of both extractors. In this initial experiment, the aggregation strategy is limited to a simple concatenation of the two feature vectors, namely those extracted from the CNN and the Action Units. This serves as a first step to evaluate the viability of the approach. Future work will explore more advanced aggregation strategies, such as weighted fusion or Principal

Component Analysis (PCA), to construct a more robust and discriminative feature representation.

The final part of the model is a set of K Gaussian mixtures, where K corresponds to the number of classes. Each mixture contains one or more Gaussian distributions $G_k = (\mu k, \Sigma k)$. To determine the optimal number of components, we rely on the Bayesian Gaussian Mixture approach (Corduneanu and Bishop, 2001), which prunes redundant components during training (Lu, 2021). Each conditional Gaussian mixture is initialized when its class appears and is trained to learn the conditional distribution of their class. Thus, each Gaussian mixture acts as an expert specialized in its own class.

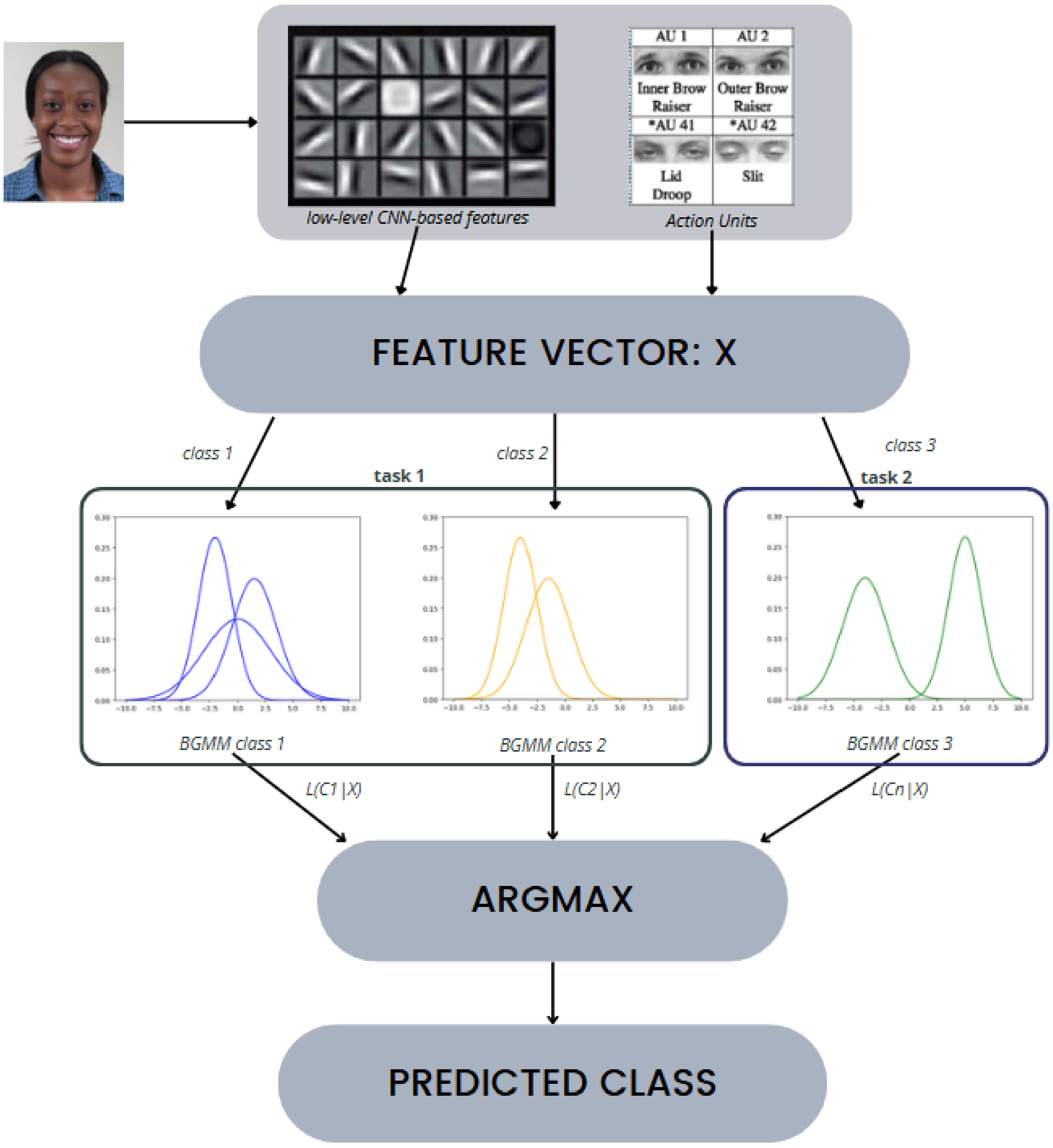

*Fig. 3. Model architecture for 2 tasks (task1=class1 + class2 and task2 = class3).*

### 2.1.3 Algorithm

During training on the initial task, the different modalities are aggregated into a feature vector that is used to train the Gaussian mixtures. For this concatenation, we adopt a basic strategy: the low-level feature vectors are normalized between 0 and 1 and then combined with the high-level features to form a single vector containing both modalities.

After aggregation, the training dataset is divided into n subsets, each corresponding to one class of the initial task. These subsets are then used to train Gaussian mixtures. Each mixture is trained with the Bayesian Gaussian Mixture approach (Corduneanu and Bishop, 2001). We observed that using diagonal covariances matrices gives the best performance when using Bayesian Gaussian Mixtures as shown in Table 1 a complete covariance matrix overfit strongly and is not able to perform with the testing set, a simple variance producing spherical representation of data shows lower performances and the best performances are achieved with a diagonal covariance matrix.

*Table 1: Accuracy of the combined features with different type of covariances, the diagonal covariances show the best performances for these features.*

| Accuracy (mean ± std) | Simple Variance | Diagonal covariance matrix | General covariance matrix |
|---|---|---|---|
| Train | 0.5838 ± 0.0087 | 0.6067± 0.003 | 1.0000 |
| Test | 0.5540 ± 0.0047 | **0.5850± 0.01** | 0.3745 |

For subsequent tasks, the same process is repeated while applying the same aggregation strategy. The Gaussian mixtures specialized in the new tasks are trained in the same way as for the initial task.

During inference, each facial image is processed by the feature extractors and aggregated following the same strategy as in training. The resulting feature vector is then passed through all Gaussian mixtures to compute the corresponding log-likelihoods. The mixture that produces the highest log-likelihood is considered the best match, and its associated class is returned as the model's prediction.

The training and inference algorithm of this architecture are presented below:

**Training algorithm**

| **Step** | **Operation** |
|---|---|
| | Input: Annotated dataset of facial expressions D = {(I, y)} |
| | Output: Set of trained class-conditional BGMMs M = {BGMM_c} |
| **1** | Initialize M ← ∅ (dictionary of class → BGMM) |
| **2** | For each task T in D: |
| **3** | If T is the first task, train CNN feature extractor Fcnn on (I, y) ∈ T |
| **4** | Initialize FeatureSet_T ← ∅ |
| **5** | For each image I in T: |
| **6** | fCNN ← Fcnn(I) (CNN feature vector) |
| **7** | fAU ← OpenFace(I) (Action Units feature vector) |
| **8** | f ← Concatenate(fCNN, fAU) (fused feature vector) |
| **9** | Add (f, label(I)) to FeatureSet_T |
| **10** | End For |
| **11** | For each class c in Classes(T): |
| **12** | Fc ← { f \| (f, y) ∈ FeatureSet_T and y = c } |
| **13** | BGMM_c ← TrainBGMM(Fc) (train class-conditional model) |
| **14** | Add (c, BGMM_c) to M |
| **15** | End For |
| **16** | End For |
| **17** | Return M |

**Inference algorithm**

| Step | Operation |
|---|---|
| | Input: Dataset of facial expressions Dtest = { I } |
| | Output: List of predicted labels Ypred |
| 1 | Initialize Ypred ← [] |
| 2 | For each image I in Dtest: |
| 3 | fCNN ← Fcnn(I) (CNN feature vector) |
| 4 | fAU ← OpenFace(I) (Action Units feature vector) |
| 5 | f ← Concatenate(fCNN, fAU) (fused feature vector) |
| 6 | Initialize Scores ← {} (dictionary class → likelihood) |
| 7 | For each (c, BGMM_c) in M: |
| 8 | logL ← BGMM_c.logLikelihood(f) |
| 9 | Scores[c] ← logL |
| 10 | End For |
| 11 | c ← argmax(Scores[c]) (class with maximum likelihood) |
| 12 | Append c to Ypred |
| 13 | End For |
| 14 | Return Ypred |

## 2.2 Experimental protocol

### 2.2.1 Continual learning

Continual learning can be separated into different types of scenarios depending on how the set of labels and the amount of data points provided during each new task vary. In this study, we focus on a common scenario called Class-Incremental Learning. In this setting, tasks are presented to the model sequentially, and the model learns each task without access to the data from previous ones. Furthermore, the labels for each task are provided to the model for training but not in the inference phase.

Formally, the training set for a task *t* can be defined as $\mathcal{D}_t = \{\mathcal{X}_t, \mathcal{Y}_t\}$ where $\mathcal{X}_t$ represent the data of a given task and $Y_t$ the corresponding labels. We assume that each task follows a distribution $\mathcal{D}_t := p(\mathcal{X}_t, \mathcal{Y}_t)$ and that there is no difference of distribution between training and testing sets. In this study, each task's data is assumed to arrive in a single batch.

Since our work focuses on facial expression classification, we divided the dataset into two parts. The initial task, used to train the feature extractor and initialize the Gaussian mixtures, consists of Ekman's basic expressions (joy, sadness, surprise, fear, anger, disgust, and neutral). To organize the remaining classes into multiple tasks, we grouped the compound expressions according to their emotional proximity. We created incremental tasks corresponding to compound variants of joy, sadness, fear, anger, and disgust. This design allows us to evaluate how our architecture handles each family of basic emotions throughout the continual learning process. Based on this principle, the incremental tasks are defined as follows:

- **Compound joy**: happily surprised, happily disgusted, awed.
- **Compound sadness**: sadly fearful, sadly angry, sadly surprised, sadly disgusted.
- **Compound fear**: fearfully angry, fearfully surprised, fearfully disgusted.
- **Compound anger**: angrily surprised, angrily disgusted, hatred.
- **Compound disgust**: disgustedly surprised, appalled.

### 2.2.2 Datasets

To enable continual learning across a wide range of emotional expressions, we require an annotated dataset containing the target categories. For this purpose, we use the Compound

Facial Expression of Emotions (CFEE) dataset. It contains 5,060 images from 230 subjects, distributed across 22 classes. The images were collected in a controlled laboratory setting: each face is centered against a white background, and each subject displays several expressions. The subjects vary in both gender and ethnicity.

In terms of labeling, the dataset includes seven basic emotion classes (including neutral) and twelve compound emotion classes such as happily-surprised or angrily–disgusted. Although compound expressions are composed of multiple basic emotions, they are not merely the mechanical combination of facial muscle activations from their components. Instead, they are considered distinct, recognizable expressions that can be reliably used for model inference.

In addition to these nineteen basic and compound expressions, the dataset also contains three so-called complex emotions: appalled, hatred, and awed. While not formally categorized as compound expressions, they can be interpreted as combinations of multiple affective states. For instance, appalled may be described as fear mixed with disgust, hatred as strong anger combined with disgust, and awed as intense joy blended with fear. These associations are derived from the semantic meaning of the original English terms.

### 2.2.3 Metrics

Continual learning performance needs to be evaluated from three perspectives: overall performance across all tasks, memory stability on previously learned tasks, and learning plasticity on new tasks (Wang et al., 2023). To address these aspects, we rely on three metrics.

The overall performance is assessed using Average Accuracy (AA) and Average Incremental Accuracy (AIA). Memory stability is measured using the Forgetting Measure (FM), which quantifies the average loss of performance on past tasks. Finally, learning plasticity is measured using the Intransigence Measure (IM), which captures the impact of learning new tasks on model performance.

Average Accuracy (AA) at task k evaluates model performance across all tasks learned so far. It is defined as the mean test accuracy on all k tasks:

$$\mathrm{AA}_k = \frac{1}{k}\sum_{j=1}^{k} a_{k,j}, \tag{1}$$

Where $a_{\mathrm{k},\mathrm{j}} \in [0,1]$ represent the accuracy on the test set of task *j* after training on task *k*.:

The Average Incremental Accuracy (AIA) represent a cumulative view of how AA evolves over time. It is defined as:

$$\mathrm{AIA}_k = \frac{1}{k}\sum_{i=1}^{k} \mathrm{AA}_i, \tag{2}$$

The stability of the model during continual learning is measured by the Forgetting Measure (FM), which evaluates how much accuracy the model has lost on previous tasks. It is computed by comparing the maximum performance of the model compared to the current performance.

For task *j* at step *k*, forgetting is computed as:

$$f_{j,k} = \max_{i \in \{1,\ldots,k-1\}} \left(a_{i,j} - a_{k,j}\right), \forall j < k. \tag{3}$$

The FM at the k-th task is defined as:

$$\mathrm{FM}_k = \frac{1}{k-1}\sum_{j=1}^{k-1} f_{j,k}. \tag{4}$$

A lower FM indicates a better ability to retain past knowledge and resist catastrophic forgetting.

The learning plasticity of the model, i.e. its ability to acquire new tasks, is evaluated using the Intransigence Measure (IM). This metric is defined as the difference between the performance of a jointly trained reference model and the performance of the continual learner:

$$\mathrm{IM}_k = a_k^* - a_{k,k} \tag{5}$$

Where $a_k^*$ represent the accuracy of randomly initialized reference model trained on the joined dataset of all the previous task $\bigcup_{j=1}^{k} D_j$.

A lower IM represents a model that is able to not lose much accuracy between its continual learning performances and its normal learning performances.

All those metrics are shown in Table 2.

*Table 2.Continual learning metrics*

| Measure | Role | Range | Optimization goal |
|---|---|---|---|
| Average Accuracy | Measure the average performance so far | [0,1] | Maximize |
| Average Incremental Accuracy | Measure the trend of improvement through the tasks | [0,1] | Maximize |
| Forgetting measure | Measure how much the old knowledge is forgotten | [0,1] | Minimize |
| Intransigence Measure | Measure difficulty in learning new tasks compared to classical training | [0,1] | Minimize |

### 2.2.4 Protocol

For this experimentation, the first set of features, referred to as the deep features, is computed using the MobileNet model (Howard et al., 2017). The architecture is modified so that the penultimate layer outputs a feature vector of 512 dimensions. The model is trained on the basic emotions using cross-entropy loss. After training the model to its best accuracy, the entire dataset is passed through the network to extract the embeddings from this layer.

The second set of features, referred to as the Action Unit (AU) features, is computed using the OpenFace tool (Baltrusaitis et al., 2018), which provides 17 Action Units representing the activation of specific facial muscles for each subject.

The two feature vectors are concatenated to generate a third feature vector, referred to as the merged features. This merged vector is the primary representation used in our experiments, although we also evaluate the two individual feature sets to compare performance across modalities.

## 2.3 Experimental results.

In our approach, the initial task of emotion recognition contains the six basic emotions. The accuracy of each feature type on these basic emotions is reported in Table 2 and shows that the performances are competitive with other models. Throughout the continual learning process, as illustrated in Figures 4, 5, and 6, accuracy decreases across all three feature representations. However, the merged feature vector combining the Action Unit features with the deep features extracted from the CNN consistently provides better performance. This concatenation yields a relative improvement of approximately 8% in classification accuracy across tasks, increasing from 0.530 with action units features alone to 0.575 with merged features, as shown in Table 3.

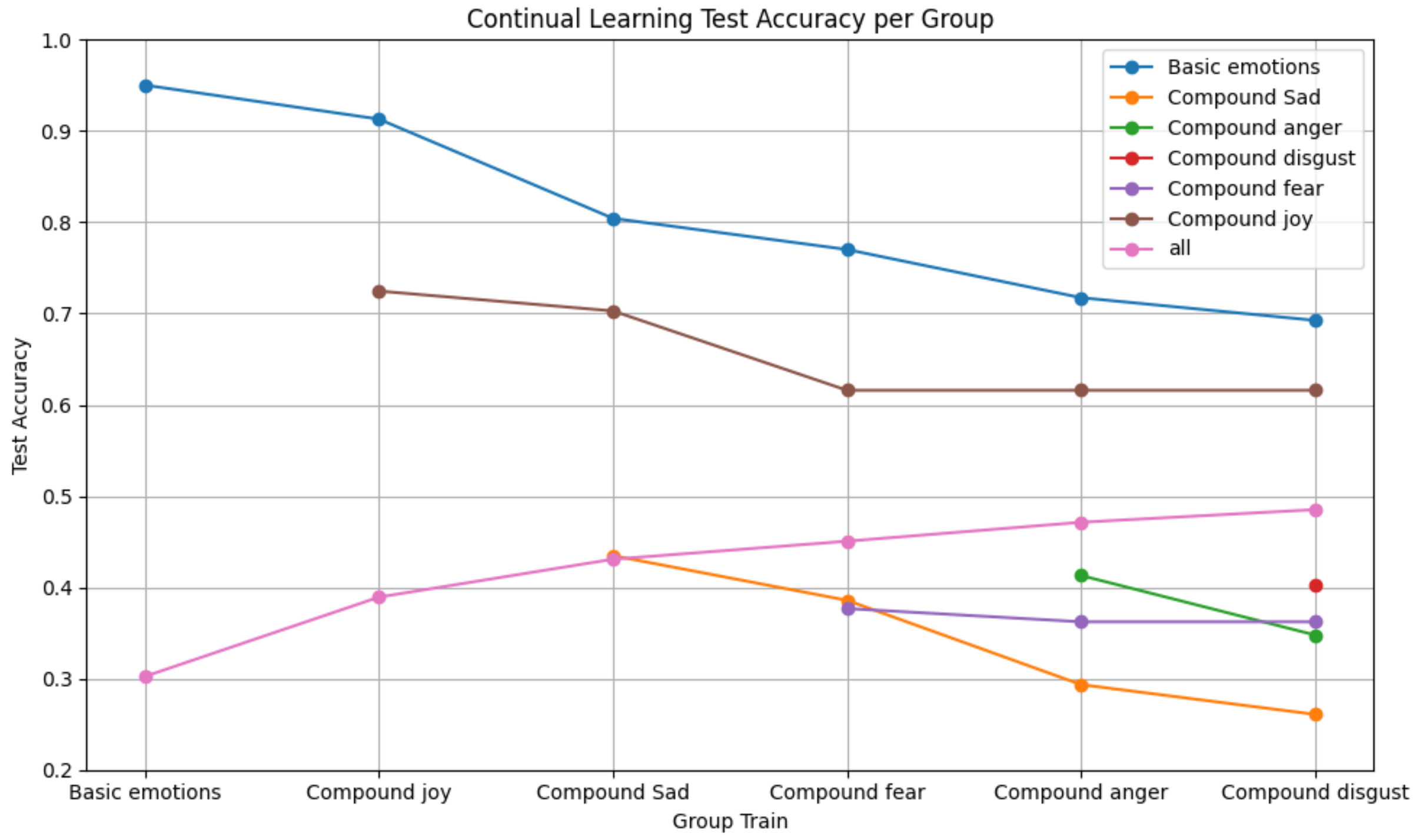

*Fig. 4. Average Accuracy of each task throughout the continual learning of the model trained on the deep feature vectors, showing the loss of accuracy on the older tasks*

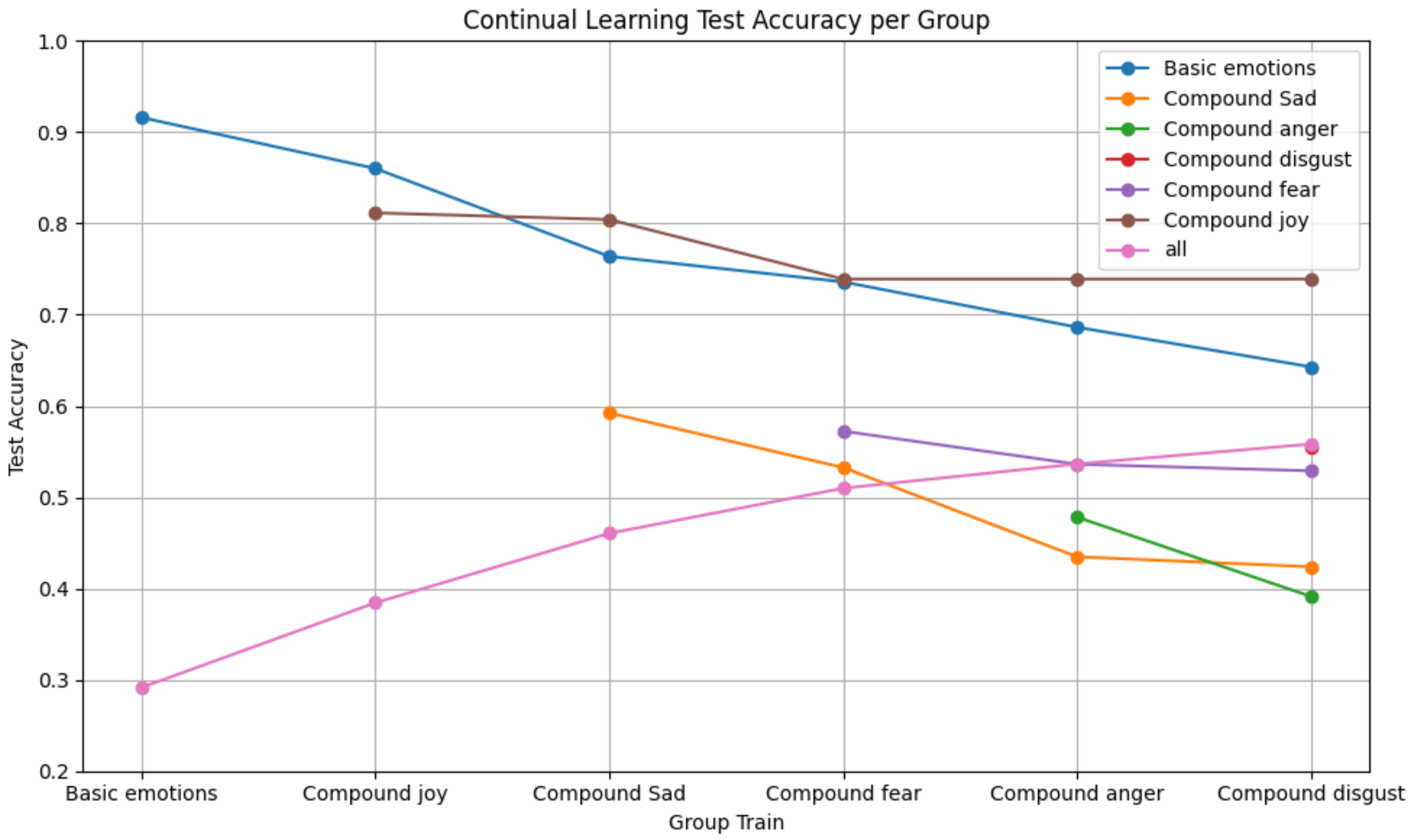

*Fig. 5. Average Accuracy of each task throughout the continual learning of the model trained on action units features, showing the loss of accuracy on the older tasks*

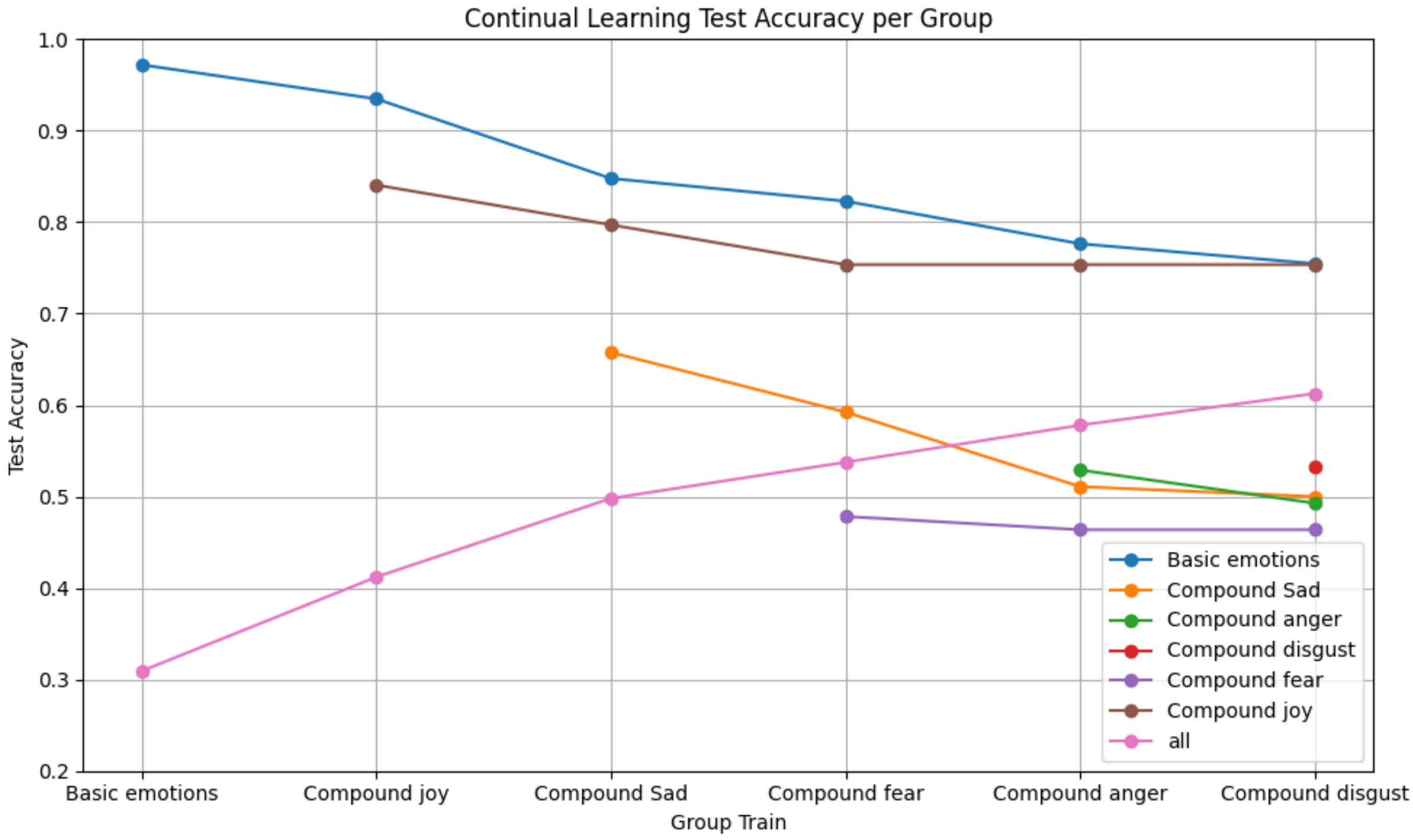

*Fig. 6. Average Accuracy (AA) of each task throughout the continual learning of the model trained on merged feature s showing the improvement of the merged features in comparison to the sole Action Units or deep features.*

*Table 3. Accuracy performances for each task at the end of the training*

| Accuracy (mean ± std) | Basic expressions | Compound joy | Compound sadness | Compound fear | Compound anger | Compound disgust | All |
|---|---|---|---|---|---|---|---|
| Deep features | 0.654±0.09 | 0.635±0.14 | 0.382±0.1 | 0.384±0.11 | 0.358±0.07 | 0.496±0.12 | 0.511±0.05 |
| AU features | 0.653±0.1 | 0.693±0.06 | 0.405±0.09 | 0.427±0.07 | 0.352±0.08 | **0.527±0.09** | 0.530±0.05 |
| Merged features | **0.744±0.02** | **0.705±0.03** | **0.467±0.03** | **0.441±0.04** | **0.378±0.03** | 0.496±0.04 | **0.575±0.01** |
| Relative evolution | +13% | **+1%** | **+15%** | **+3%** | +5% | **-5%** | **+8%** |

Figures 7, 8, and 9 present the performance of the model using three incremental metrics: Average Incremental Accuracy, Forgetting Measure, and Intransigence Measure. These metrics respectively capture the evolution of accuracy during continual learning, the stability of the model in retaining knowledge from previous tasks, and the ability of the model to acquire new information during training. Across all three graphs, the concatenation of deep features and Action Unit features consistently yields the best results.

In Figure 7, we report both the Average Accuracy and Average Incremental Accuracy of the three feature representations. The concatenated features show clear improvements in both metrics. While the deep features achieve strong performance on the initial task, their accuracy declines steadily during continual learning. This behavior is expected, since the deep features were extracted from a model trained exclusively on the initial task and thus generalize poorly to new tasks. In contrast, the Action Unit features start with lower performance on the initial task but degrade less across incremental tasks. The combined features, however, capture the strengths of both representations: they perform well on the initial task and retain stronger performance throughout continual learning. This demonstrates that the BGMM can effectively leverage the merged feature vector to infer the correct labels, even when the features are not produced by a model specialized for those classes.

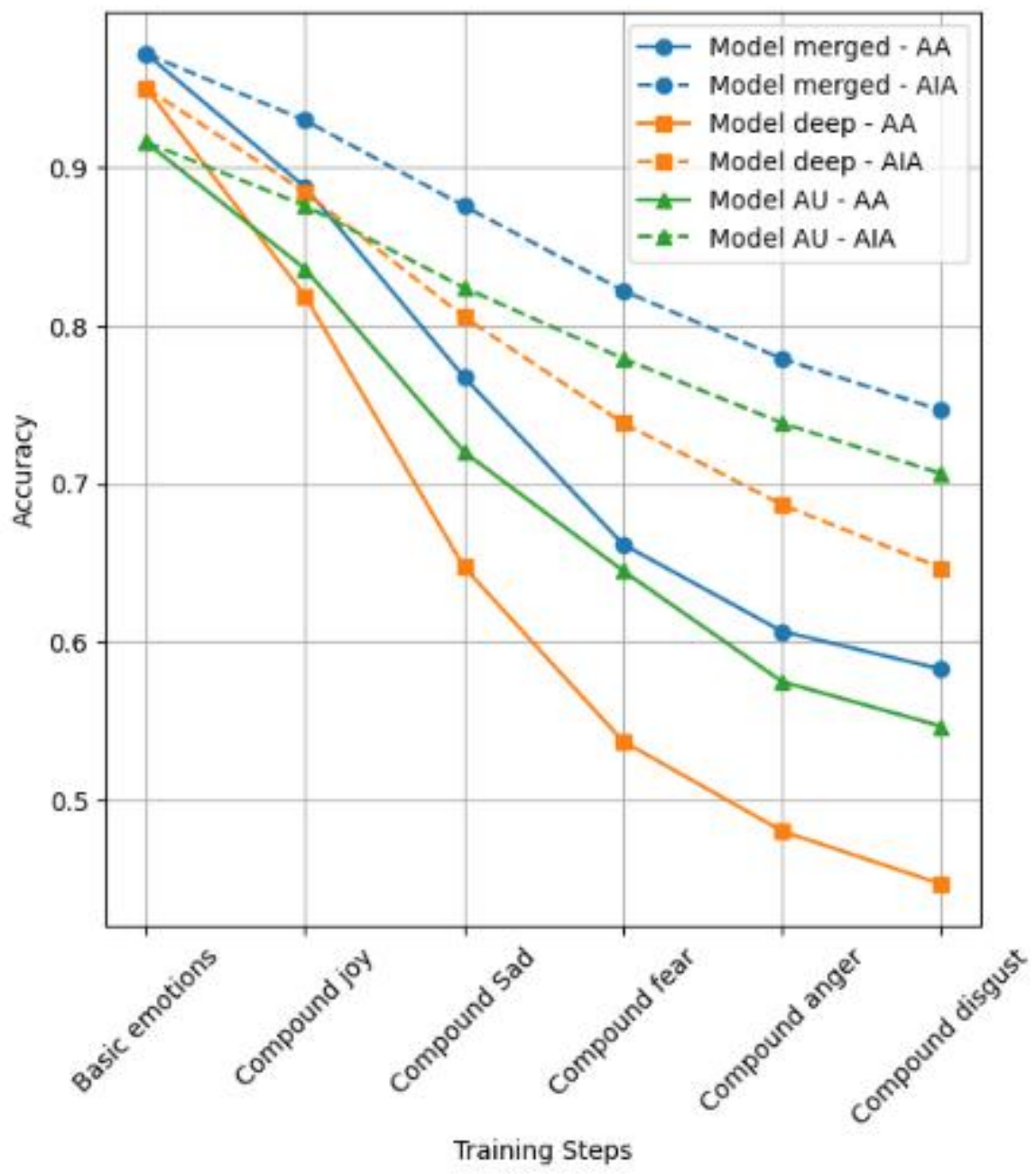

*Fig. 7. Average Accuracy (AA) and Average Incremental Accuracy (AIA) throughout the continual learning process for the three feature vectors. Showing the best performances for the merged features in Average Accuracy and Average Incremental Accuracy.*

Figure 8 shows the average forgetting of our approach on the three feature vectors. We observe lower forgetting with the concatenated feature vector, which indicates less performance loss on previous tasks during the continual learning process. The purpose of this metric is to highlight the stability of the proposed architecture. We can see that the merged features start with the same level of forgetting as the deep features on the first incremental task but gradually forget less than the two individual feature vectors as training progresses. This demonstrates that the combined features are more resilient than the separate vectors throughout the continual learning process.

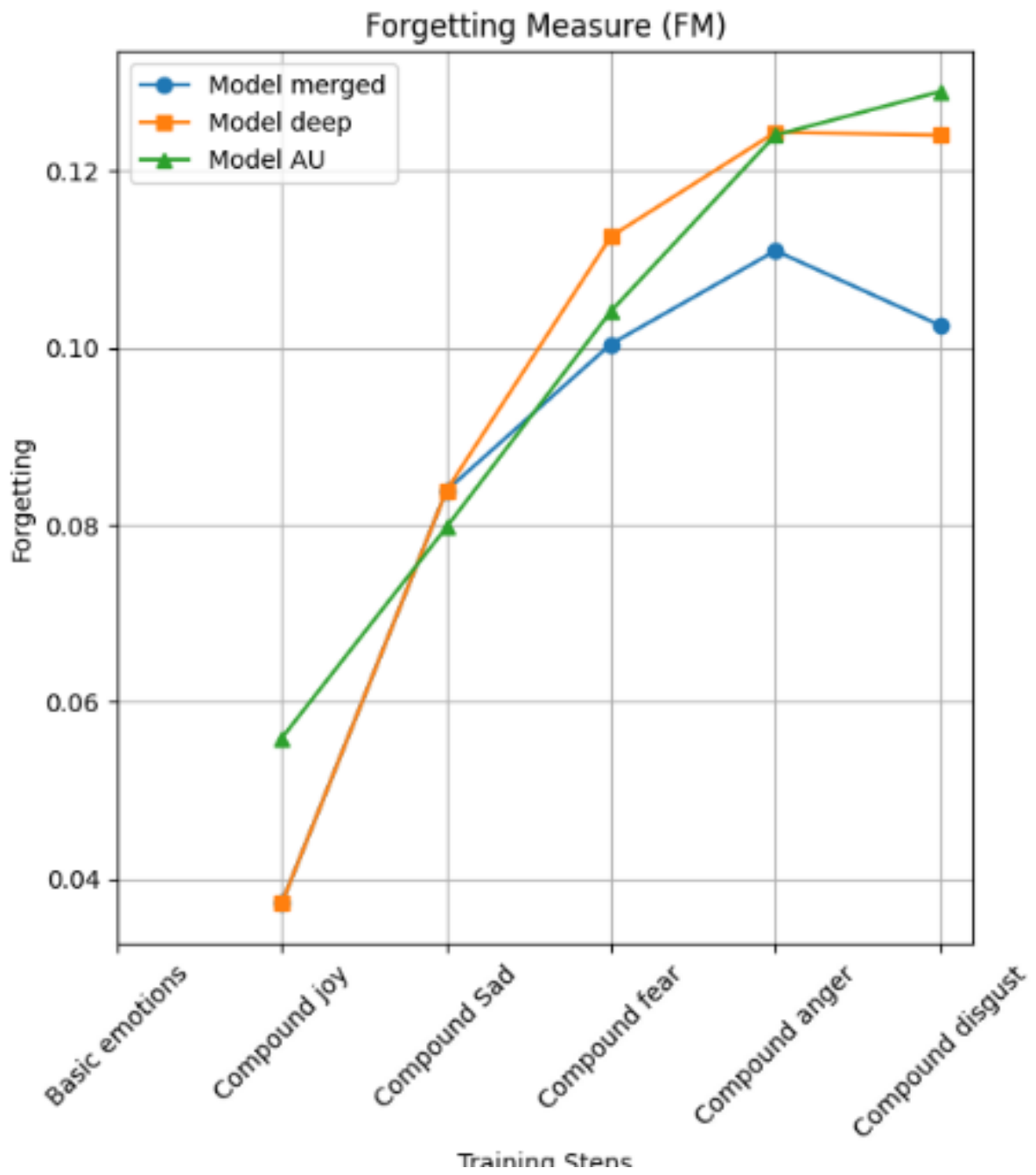

*Fig. 8. Forgetting Measure (FM) across the continual learning process for the three feature vectors. The results indicate that the merged feature representation mitigates forgetting more effectively than individual feature types.*

Figure 9 shows the intransigence performance of the model during continual learning across different tasks. Since our approach learns each class probability distribution separately, it does not suffer from accuracy loss due to the continual learning process. As a result, the Intransigence Measure remains equal to or close to zero for all feature vectors. This reflects the ability of the proposed architecture to maintain separate representations for each class. Consequently, the number of tasks and their order do not influence the performance of the architecture. The only task with a strong impact on overall performance is the initial task, because the CNN used for deep feature extraction is trained on this smaller set of labels. A reduced label set lowers the quality of the extracted feature vector and thus affects the full model performance.

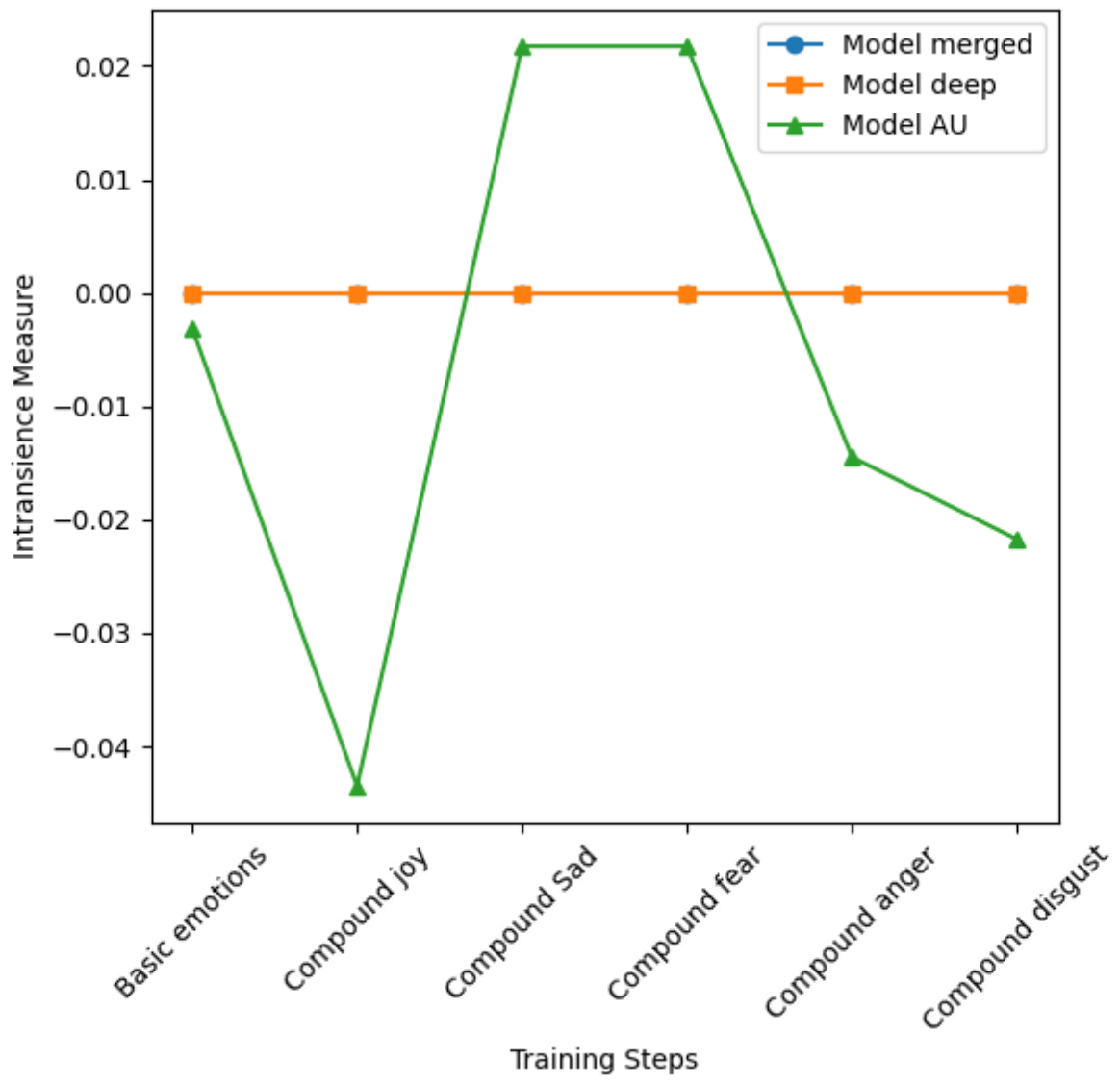

*Fig. 9. Intransigence Measure (IM) across the continual learning process for the three feature vectors, indicating that the addition of new tasks has a negligible effect on the model's performance.*

For each incremental metric, our approach achieves better performance than the individual feature vectors. The combined feature representation is able to compensate for the weaknesses of each separate modality. The proposed architecture suffers less from catastrophic forgetting, maintains higher accuracy, and does not experience performance degradation due to the continual learning setting. These results demonstrate the capability of our approach to effectively address the continual learning problem in Facial Expression Recognition.

# 3 Discussion

Each feature type exhibits different strengths and weaknesses. Figure 1 illustrates which features perform better for each expression. In the last row, negative values correspond to classes where the Action Unit features predicted the correct label while the deep features did not, and positive values indicate the opposite case. We observe that the deep features perform better on the basic expressions of the initial task but show weaker results on compound expressions. In contrast, the Action Unit features generally perform better on compound expressions, apart from the “appalled” class.

This outcome is consistent with expectations: deep CNN features, trained specifically on basic expressions, achieve higher accuracy on those categories but generalize poorly to compound ones. Conversely, Action Unit vectors, which encode facial muscle activations, demonstrate more stable performance across tasks.

Result on test by class

| | neutral | happy | sad | fear | angry | surprise | disgust | Happily surprised | Happily disgusted | Sadly fearful | Sadly angry | Sadly surprised | Sadly disgusted | Fearfully angry | Fearfully surprised | Fearfully disgusted | Angrily surprised | Angrily disgusted | Disgustedly surprised | Appalled | Hatred | Awed |
|---|---|---|---|---|---|---|---|---|---|---|---|---|---|---|---|---|---|---|---|---|---|---|
| Full Features | 45 | 44 | 40 | 11 | 31 | 37 | 22 | 44 | 42 | 26 | 23 | 30 | 17 | 22 | 25 | 28 | 32 | 15 | 34 | 14 | 15 | 25 |
| Deep Features | 45 | 41 | 38 | 11 | 31 | 39 | 26 | 37 | 35 | 23 | 24 | 24 | 13 | 17 | 19 | 19 | 31 | 6 | 30 | 15 | 13 | 22 |
| AU Features | 39 | 40 | 24 | 16 | 23 | 27 | 21 | 45 | 39 | 20 | 23 | 26 | 14 | 23 | 27 | 26 | 28 | 23 | 27 | 1 | 18 | 21 |
| Deep - AU Features | 6 | 1 | 14 | -5 | 8 | 12 | 5 | -8 | -4 | 3 | 1 | -2 | -1 | -6 | -8 | -7 | 3 | -17 | 3 | 14 | -5 | 1 |

*Fig. 10. Difference in correctly recognized test samples between the deep feature model and the AU model on the CFEE dataset, highlighting that deep features perform better on basic expressions, while AU features show superior performance on compound expressions.*

We also evaluated how our approach compares to an idealized scenario in which, between the two feature sets, the model always selects the best-performing one. To simulate this, predictions were generated using both models, and a sample was considered correct if either of the feature-based models produced the right label. As shown in Table 4, while the merged features outperform each feature set individually, they still fall short of the performance achieved by this best potential baseline. This gap is likely due to the simplicity of the concatenation strategy used in our current implementation.

*Table 4. Final and potential best accuracy of the model through different features*

| Metric | Merged features | Deep features | Action unit features | Potential best features |
|---|---|---|---|---|
| Test Accuracy (mean ± std) | 0.614±0.01 | 0.552±0.02 | 0.545±0.04 | 0.738±0.03 |

# 4 Limitations and future work

Our current architecture has several limitations. First, it relies on a large initial task to train the CNN feature extractor. This dependency reduces the flexibility of the continual learning analysis, since performance in later tasks is strongly influenced by the first training step. In future work, we plan to explore other types of feature extractors and descriptors, such as Histogram of Oriented Gradients (HOG) or Local Binary Patterns (LBP), to reduce this reliance and diversify the extracted representations.

A second limitation is the feature fusion strategy, which is currently restricted to simple concatenation. While this provides an initial proof of concept, it does not fully exploit the complementarity between modalities. Future work will therefore investigate more advanced fusion methods, such as dimensionality reduction (e.g., PCA) or dynamic weighting mechanisms, to better emphasize the most informative components of the feature vectors. These strategies may further improve performance, particularly for continually learning on complex compound expressions.

Finally, our experiments were conducted on the CFEE dataset, which contains numerous labels and was collected under control conditions, making training and evaluation easier. However, this dataset does not capture the variability of real-world scenarios. As a next step, we plan to extend our experiments to more challenging datasets such as RAF-DB, which include greater diversity in pose, illumination, and occlusion, thereby allowing us to evaluate our approach under more realistic conditions.

# 5 Conclusion

In this work, we introduced a modular architecture for facial expression recognition (FER) in a continual learning setting. Our approach combines deep features extracted from a CNN with Action Units (AUs) to build a hybrid representation of facial expressions. This is then modeled using an ensemble of Bayesian Gaussian Mixture Models with each one conditionally learnt on an emotion class. This architecture allows the system to incrementally acquire new emotional categories including compound expressions, while effectively mitigating catastrophic forgetting which is a core challenge in continual learning.

Through experiments conducted on expression-specific datasets Compound Facial Expression of Emotion (CFEE), we showed that aggregating different feature types leads to more accurate performance during the continual learning process. The merged feature consistently outperformed models based on individual feature types across all evaluated metrics. While our current concatenation strategy provides encouraging results, it does not yet match the potential upper bound where the most informative modality is always selected.

This work contributes to the development of adaptive and emotionally intelligent systems that can evolve without the need for retraining on all past data. In future work, we aim to explore more advanced feature aggregation mechanisms to further enhance performance in dynamic and unconstrained real-world environments.

# 6 References

Abhishree, T.M., Latha, J., Manikantan, K., Ramachandran, S., 2015. Face Recognition Using Gabor Filter Based Feature Extraction with Anisotropic Diffusion as a Pre-processing Technique. Procedia Comput. Sci., International Conference on Advanced Computing Technologies and Applications (ICACTA) 45, 312–321. https://doi.org/10.1016/j.procs.2015.03.149

Alzubaidi, L., Zhang, J., Humaidi, A.J., Al-Dujaili, A., Duan, Y., Al-Shamma, O., Santamaría, J., Fadhel, M.A., Al-Amidie, M., Farhan, L., 2021. Review of deep learning: concepts, CNN architectures, challenges, applications, future directions. J. Big Data 8, 53. https://doi.org/10.1186/s40537-021-00444-8

Baltrusaitis, T., Zadeh, A., Lim, Y.C., Morency, L.-P., 2018. OpenFace 2.0: Facial Behavior Analysis Toolkit, in: 2018 13th IEEE International Conference on Automatic Face & Gesture Recognition (FG 2018). Presented at the 2018 13th IEEE International

Conference on Automatic Face & Gesture Recognition (FG 2018), IEEE, Xi'an, pp. 59–66. https://doi.org/10.1109/FG.2018.00019
Castellano, G., Kessous, L., Caridakis, G., 2008. Emotion Recognition through Multiple Modalities: Face, Body Gesture, Speech, in: Peter, C., Beale, R. (Eds.), Affect and Emotion in Human-Computer Interaction, Lecture Notes in Computer Science. Springer Berlin Heidelberg, Berlin, Heidelberg, pp. 92–103. https://doi.org/10.1007/978-3-540-85099-1_8
Chaudhry, A., Ranzato, M., Rohrbach, M., Elhoseiny, M., 2019. Efficient Lifelong Learning with A-GEM. https://doi.org/10.48550/arXiv.1812.00420
Cootes, T.F., Taylor, C.J., Cooper, D.H., Graham, J., 1995. Active Shape Models-Their Training and Application. Comput. Vis. Image Underst. 61, 38–59. https://doi.org/10.1006/cviu.1995.1004
Corduneanu, A., Bishop, C., 2001. Variational Bayesian model selection for mixture distribution. Artif. Intell. Stat. 18, 27–34.
Cowie, R., Douglas-Cowie, E., Tsapatsoulis, N., Votsis, G., Kollias, S., Fellenz, W., Taylor, J.G., 2001. Emotion recognition in human-computer interaction. IEEE Signal Process. Mag. 18, 32–80. https://doi.org/10.1109/79.911197
Dempster, A.P., Laird, N.M., Rubin, D.B., 1977. Maximum Likelihood from Incomplete Data Via the *EM* Algorithm. J. R. Stat. Soc. Ser. B Stat. Methodol. 39, 1–22. https://doi.org/10.1111/j.2517-6161.1977.tb01600.x
Deng, J., Dong, W., Socher, R., Li, L.-J., Li, K., Fei-Fei, L., 2009. ImageNet: A large-scale hierarchical image database, in: 2009 IEEE Conference on Computer Vision and Pattern Recognition. Presented at the 2009 IEEE Conference on Computer Vision and Pattern Recognition, pp. 248–255. https://doi.org/10.1109/CVPR.2009.5206848
Du, S., Tao, Y., Martinez, A.M., 2014. Compound facial expressions of emotion. Proc. Natl. Acad. Sci. U. S. A. 111. https://doi.org/10.1073/pnas.1322355111
Ekman, P., Friesen, W.V., 1978. Facial action coding system. Environ. Psychol. Nonverbal Behav.
Ekman, P., Friesen, W.V., 1971. Constants across cultures in the face and emotion. J. Pers. Soc. Psychol. 17, 124–129. https://doi.org/10.1037/h0030377
French, R.M., 1999. Catastrophic forgetting in connectionist networks. Trends Cogn. Sci. 3, 128–135. https://doi.org/10.1016/S1364-6613(99)01294-2
Geslin, E., 2013. Process of inducing emotions in virtual environments and video games.
Goodfellow, I.J., Erhan, D., Carrier, P.L., Courville, A., Mirza, M., Hamner, B., Cukierski, W., Tang, Y., Thaler, D., Lee, D.-H., Zhou, Y., Ramaiah, C., Feng, F., Li, R., Wang, X., Athanasakis, D., Shawe-Taylor, J., Milakov, M., Park, J., Ionescu, R., Popescu, M., Grozea, C., Bergstra, J., Xie, J., Romaszko, L., Xu, B., Chuang, Z., Bengio, Y., 2013. Challenges in Representation Learning: A Report on Three Machine Learning Contests, in: Lee, M., Hirose, A., Hou, Z.-G., Kil, R.M. (Eds.), Neural Information Processing. Springer, Berlin, Heidelberg, pp. 117–124. https://doi.org/10.1007/978-3-642-42051-1_16
Graesser, A.C., D'Mello, S., 2012. Emotions During the Learning of Difficult Material, Psychology of Learning and Motivation - Advances in Research and Theory. https://doi.org/10.1016/B978-0-12-394293-7.00005-4
Hochreiter, S., Schmidhuber, J., 1997. Long Short-Term Memory. Neural Comput. 9, 1735–1780. https://doi.org/10.1162/neco.1997.9.8.1735
Howard, A.G., Zhu, M., Chen, B., Kalenichenko, D., Wang, W., Weyand, T., Andreetto, M., Adam, H., 2017. MobileNets: Efficient Convolutional Neural Networks for Mobile Vision Applications [WWW Document]. arXiv.org. URL https://arxiv.org/abs/1704.04861v1 (accessed 6.19.25).

Kemker, R., Kanan, C., 2018. FearNet: Brain-inspired model for incremental learning. 6th Int. Conf. Learn. Represent. ICLR 2018 - Conf. Track Proc. 1–16.

Kirkpatrick, J., Pascanu, R., Rabinowitz, N., Veness, J., Desjardins, G., Rusu, A.A., Milan, K., Quan, J., Ramalho, T., Grabska-Barwinska, A., Hassabis, D., Clopath, C., Kumaran, D., Hadsell, R., 2017. Overcoming catastrophic forgetting in neural networks. Proc. Natl. Acad. Sci. 114, 3521–3526. https://doi.org/10.1073/pnas.1611835114

Kopalidis, T., Solachidis, V., Vretos, N., Daras, P., 2024. Advances in Facial Expression Recognition: A Survey of Methods, Benchmarks, Models, and Datasets. Information 15, 135. https://doi.org/10.3390/info15030135

Krizhevsky, A., Hinton, G., 2009. Learning multiple layers of features from tiny images.

Li, S., Deng, W., Du, J., 2017. Reliable Crowdsourcing and Deep Locality-Preserving Learning for Expression Recognition in the Wild, in: 2017 IEEE Conference on Computer Vision and Pattern Recognition (CVPR). Presented at the 2017 IEEE Conference on Computer Vision and Pattern Recognition (CVPR), IEEE, Honolulu, HI, pp. 2584–2593. https://doi.org/10.1109/CVPR.2017.277

Lopez-Paz, D., Ranzato, M., 2017. Gradient Episodic Memory for Continual Learning. https://doi.org/10.48550/arXiv.1706.08840

Lu, J., 2021. A survey on Bayesian inference for Gaussian mixture model. https://doi.org/10.48550/arXiv.2108.11753

Lucey, P., Cohn, J.F., Kanade, T., Saragih, J., Ambadar, Z., Matthews, I., Ave, F., 2010. The Extended Cohn-Kanade Dataset ( CK + ): A complete dataset for action unit and emotion-specified expression 94–101.

Pham, Q., Liu, C., Hoi, S., 2021. DualNet: Continual Learning, Fast and Slow. https://doi.org/10.48550/arXiv.2110.00175

Picard, R.W., 2000. Affective computing. MIT press.

Ramesh, R., Chaudhari, P., 2022. Model Zoo: A Growing “Brain” That Learns Continually. https://doi.org/10.48550/arXiv.2106.03027

Robins, A., 1995. Catastrophic Forgetting, Rehearsal and Pseudorehearsal. Connect. Sci. 7, 123–146. https://doi.org/10.1080/09540099550039318

Tariq, U., Yang, J., Huang, T.S., 2013. Maximum margin GMM learning for facial expression recognition, in: 2013 10th IEEE International Conference and Workshops on Automatic Face and Gesture Recognition (FG). Presented at the 2013 10th IEEE International Conference and Workshops on Automatic Face and Gesture Recognition (FG), pp. 1–6. https://doi.org/10.1109/FG.2013.6553794

Wang, L., Zhang, X., Su, H., Zhu, J., 2024. A Comprehensive Survey of Continual Learning: Theory, Method and Application. https://doi.org/10.48550/arXiv.2302.00487

Wang, L., Zhang, X., Su, H., Zhu, J., 2023. A Comprehensive Survey of Continual Learning: Theory, Method and Application.

Zhang, Z., Lyons, M., Schuster, M., Akamatsu, S., 1998. Comparison between Geometry-Based and Gabor Wavelets-Based Facial Expression Recognition Using Multi-Layer Perceptron, textordmasculine Proc. Int’l Conf. Automatic Face and Gesture Recognition. https://doi.org/10.1109/AFGR.1998.670990